\documentclass[]{jingdong}
\usepackage[utf8]{inputenc}
\usepackage[toc,page,header]{appendix}

\usepackage{latexsym}
\usepackage[T1]{fontenc}
\usepackage{inconsolata}
\usepackage{textcomp}
\usepackage{xkeyval}
\usepackage{amsfonts}
\usepackage{nicefrac}
\usepackage{ifthen}
\usepackage{stackengine}
\usepackage{listofitems}
\usepackage{calc}
\usepackage{colortbl}
\usepackage{amssymb}
\usepackage{amsthm}
\usepackage{mathrsfs}
\usepackage{pifont}
\usepackage{MnSymbol}
\usepackage{enumitem}
\usepackage{amsmath}
\usepackage{mathtools}

\makeatletter
\DeclareRobustCommand\onedot{\futurelet\@let@token\@onedot}
\def\@onedot{\ifx\@let@token.\else.\null\fi}

\makeatother

\usepackage{cleveref}
\theoremstyle{plain}

\theoremstyle{definition}

\theoremstyle{remark}

\usepackage[textsize=tiny]{todonotes}

\usepackage{tabularx}
\usepackage{xurl}
\usepackage{tikz}
\usetikzlibrary{arrows.meta,positioning,shapes.geometric}

\newcolumntype{Y}{>{\raggedright\arraybackslash}X}

\newcommand{\pie}{Play In Editor (PIE)}
\newcommand{\mrq}{Movie Render Queue (MRQ)}

\title{Building Pretraining Data for World Models:\\
An Unreal Engine-Based Pipeline for Action-Conditioned Video Generation}
\author[1,2]{\normalsize{Haoyu Wang}}
\author[1,3]{\normalsize{Songchun Zhang}}
\author[1]{\normalsize{Haoran Li}}
\author[1]{\normalsize{Haoyang Huang}}
\author[1]{\normalsize{Zeyue Xue}}
\author[1]{\normalsize{Nan Duan}}

\affiliation{\small{$^1$Joy Future Academy, JD}}
\affiliation{\small{$^2$Tsinghua University}}
\affiliation{\small{$^3$The Hong Kong University of Science and Technology}}

\abstract{
Action-conditioned video models require large-scale visual data paired with control signals that are temporally aligned with the resulting scene transitions. Such supervision is difficult to obtain from ordinary real-world video because the actions that caused each visual change are typically unknown. We present a large-scale synthetic data production pipeline built on Unreal Engine for generating action-conditioned, multi-view video. To accommodate the different execution requirements of real-time physics and high-quality offline rendering, the pipeline executes trajectory generation and final rendering in two stages: Stage I runs real physics in PIE and records per-frame character states, control inputs, and camera states into an intermediate trajectory representation; Stage II replays those trajectories in a new engine process and renders them offline with Movie Render Queue (MRQ)\@. Around this core, we develop a distributed production system with cache-aware task partitioning, node-local slot scheduling, automated scene screening, aesthetic and luminance filtering, partial-output recovery, asynchronous upload, and continuous cluster health monitoring. The production cluster contains 25 servers with eight NVIDIA RTX 5090 GPUs per server. From 2,384 asset packs, 429 levels were retained for production together with a pool of 40 humanoid characters. The pipeline has produced 2,691 hours of 1080p video and 6,076 hours of 720p video. We describe the system architecture, the implementation decisions that emerged from production failures, and the limitations of using perceptual quality proxies for world-model data curation. The pipeline described in this report constitutes the Unreal Engine synthetic-data production component used in EchoWM.
}

\date{\today}
\checkdata[Project Page (EchoWM)]{\url{https://echo-team-joy-future-academy-jd.github.io/Echo-1.5-Page/wm/}}

\begin{document}
\maketitle

\section{Introduction}

Recent video world models are increasingly moving toward interactive generation, where future visual observations are conditioned on user or agent actions. Systems such as Genie, GameNGen, WorldPlay, Matrix-Game 3.0, SANA-WM, DreamX-World, and EchoWM have demonstrated increasingly high-resolution, long-horizon, and controllable environment generation \citep{bruce2024genie,valevski2024gamengen,sun2025worldplay, wang2026matrixgame3,zhu2026sanawm,dreamx2026world,zhang2026echowm}. This progress places growing demands on the underlying training data: in addition to diverse visual observations, interactive world models require reliable temporal alignment between video, actions, and often camera motion.

Constructing such data at scale remains non-trivial. Internet video provides broad visual diversity, but actions and camera motion are rarely observed directly and must instead be inferred or reconstructed. Gameplay data provides explicit controls, but is typically tied to a particular game or environment. Recent world-model systems therefore increasingly employ heterogeneous data engines that combine gameplay, Internet video, and synthetic 3D environments. For example, Matrix-Game 3.0, DreamX-World, and EchoWM all incorporate synthetically generated or simulated data as part of a larger data construction pipeline \citep{wang2026matrixgame3,dreamx2026world,zhang2026echowm}.

This report focuses specifically on the synthetic-data production component of this broader ecosystem. Our goal is to convert a large collection of heterogeneous third-party Unreal Engine assets into a scalable source of high-quality, multi-view video with precisely aligned control signals. In this setting, actions are generated by the pipeline rather than inferred after video collection, while character states and camera poses can be recorded directly from the engine.

Building such a pipeline on top of Unreal Engine introduces several practical constraints. In particular, physically valid character motion relies on the engine's runtime execution so that collision, gravity, and other physical interactions can be resolved, whereas high-quality video generation relies on Movie Render Queue (MRQ), whose offline rendering workflow is not naturally compatible with running the same physics simulation in real time. Under these constraints, we implement production in two stages: trajectories are first executed and recorded in Play In Editor (PIE), and the recorded character and camera states are subsequently replayed in a separate Unreal Engine process for MRQ rendering. The two-stage workflow is therefore a practical consequence of the engine's execution model rather than an independent modeling assumption.

Scaling this procedure from an individual scene to a production cluster introduces additional systems requirements. Third-party Unreal assets vary substantially in scene organization, collision geometry, built-in cinematics, texture-streaming behavior, and renderability. Repeated rendering also incurs scene-specific shader and texture initialization costs, making cache locality important for scheduling. Finally, continuously running hundreds of graphics processes requires mechanisms for partial-output recovery, asynchronous uploading, storage management, and automatic failure detection. Rather than treating these issues as isolated implementation details, our pipeline incorporates asset curation, cache-aware distributed scheduling, quality filtering, and fault recovery into a unified production workflow.

The resulting system runs on 25 compute nodes with eight NVIDIA RTX 5090 GPUs per node. From 2,384 Fab asset packs, we retain 429 production levels after renderability and visual-quality screening, and combine them with a pool of 40 characters. The pipeline has produced 2,691 hours of 1080p video and 6,076 hours of 720p video, with five synchronized camera views and frame-aligned action and pose metadata.

Our pipeline makes the following contributions:
\begin{enumerate}[leftmargin=1.6em]
    \item \textbf{A production pipeline for precisely aligned action-conditioned video data.}
    The system generates synchronized multi-view video together with frame-level actions, character states, and camera poses from heterogeneous Unreal Engine environments.
    \item \textbf{A scalable execution architecture for large-scale Unreal Engine production.}
    The system combines persistent node-level scene pools, node-local rendering slots, cache-aware scheduling, and multi-process GPU execution to sustain production across a 200-GPU cluster.
    \item \textbf{An automated curation and reliability pipeline} covering scene screening, aesthetic and luminance filtering, partial-output recovery, asynchronous upload, failure snapshots, and cluster health monitoring.
    \item \textbf{A large-scale production effort} spanning 429 environments, 40 humanoid characters, five synchronized camera views, and thousands of hours of generated action-conditioned video.
\end{enumerate}

This work focuses on the infrastructure for constructing action-conditioned video data rather than introducing a new world-model architecture. The pipeline described in this report constitutes the Unreal Engine synthetic-data production component used in EchoWM~\cite{zhang2026echowm}.

\section{Related Work}

\subsection{Interactive Generative World Models}

Recent video-based world models have rapidly progressed toward high-resolution, long-horizon, and interactive generation. LingBot-World extends open-world simulation to minute-scale generation with real-time interaction \citep{gao2026lingbotworld}, while WorldPlay focuses on streaming generation with long-term geometric consistency \citep{sun2025worldplay}. Matrix-Game 3.0 further combines 720p real-time generation with camera-aware long-horizon memory \citep{wang2026matrixgame3}. SANA-WM targets efficient minute-scale generation with precise metric 6-DoF camera control \citep{zhu2026sanawm}, and DreamX-World supports controllable navigation, scene revisiting, and promptable events across photorealistic, game-style, and stylized environments \citep{dreamx2026world}. Recent omnimodal systems such as Cosmos 3 and EchoWM further extend world modeling beyond visual prediction by jointly modeling action and additional modalities \citep{nvidia2026cosmos3,zhang2026echowm}.

Collectively, these systems demonstrate a shift from domain-specific visual prediction toward general-purpose interactive video generation. As model capacity and interaction horizons increase, however, their training increasingly depends on large-scale video data with reliable camera geometry, actions, and temporal alignment. Our work focuses on this upstream data production problem rather than on the predictive model itself.

\subsection{Data Construction for Interactive World Models}

Recent world models increasingly rely on heterogeneous data engines that combine synthetic environments, gameplay, and Internet video. Matrix-Game 3.0 constructs an industrial-scale data engine integrating Unreal Engine synthetic scenes, automated AAA-game collection, and real-world video augmentation to produce Video--Pose--Action--Prompt tuples \citep{wang2026matrixgame3}. DreamX-World similarly combines camera-accurate Unreal Engine rendering, action-rich gameplay recordings, and real-world video with recovered camera geometry \citep{dreamx2026world}. SANA-WM instead derives metric-scale 6-DoF camera supervision from public video through a dedicated annotation pipeline \citep{zhu2026sanawm}. Beyond interactive navigation, Qwen-RobotWorld and GigaWorld-0 scale world-model data construction toward embodied intelligence and downstream policy learning
\citep{zhang2026qwenrobotworld,gigaworld2025}.

These works make clear that data construction has become a first-class component of modern world-model systems. Nevertheless, the synthetic-data portion of such pipelines is typically described only as one component of a larger model-training system. This report instead focuses specifically on the production infrastructure required to turn a large collection of heterogeneous Unreal Engine assets into high-quality, multi-view, action-aligned video at cluster scale, including asset screening, trajectory generation, offline rendering, distributed scheduling, quality control, and failure recovery.
Another relevant direction is the construction and curation of large video corpora with richer spatial annotations. SpatialVID processes a large collection of in-the-wild videos into millions of clips and augments them with camera poses, depth, dynamic masks, captions, and motion instructions through a hierarchical filtering and annotation pipeline~\citep{wang2025spatialvid}. Cosmos likewise treats large-scale video curation as a first-class component of world-foundation-model development~\citep{nvidia2025cosmos}. These efforts highlight that model performance depends not only on model architecture but also on systematic data filtering, annotation, and coverage.

Our pipeline shares this data-centric perspective but operates in a fully synthetic control setting. Spatial quantities and actions need not be estimated from the rendered video because they are available from the engine during generation. Conversely, the use of third-party synthetic scenes introduces its own distributional biases and requires explicit screening for collision validity, renderability, exposure, texture quality, and scene semantics. The aesthetic and luminance filters used in our current production pipeline are adapted from the SpatialVID scoring tools, while the broader question of how such perceptual filters correlate with downstream world-model utility remains open.

\section{Data Generation Overview}

\begin{figure}[h]
\centering
\includegraphics[width=0.9\linewidth]{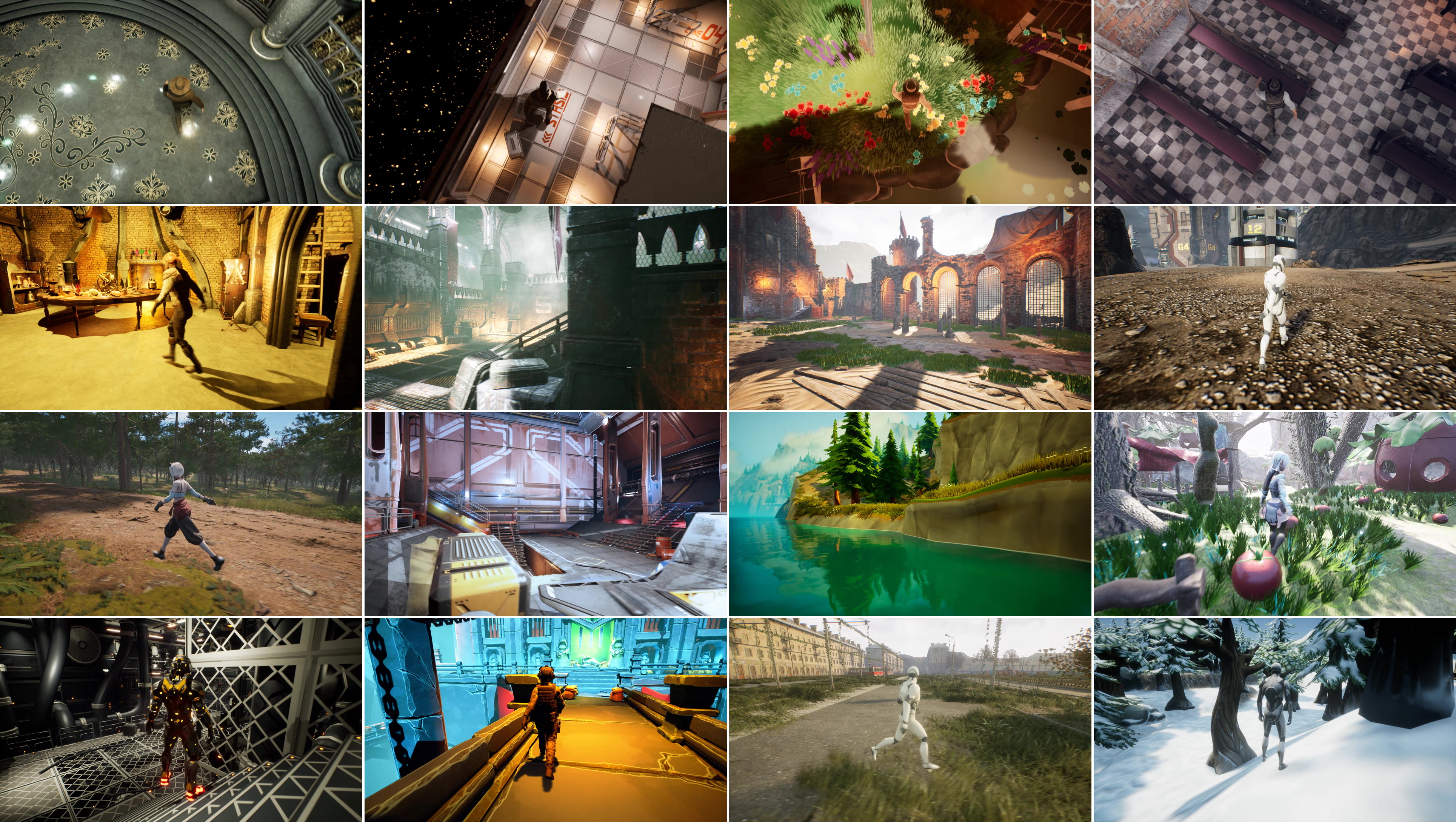}
\caption{
    Representative rendered observations from the five camera configurations used in the dataset. The top row shows \textit{top-down} views, the bottom row shows \textit{back} views, the center four examples show \textit{first-person} views, and the examples on the left and right sides show the corresponding \textit{left-} and \textit{right-side} views. Different environments are intentionally sampled to illustrate both viewpoint variation and scene diversity.
}
\label{fig:data-examples}
\end{figure}

\subsection{Action-Conditioned Multi-View Data}
Each generated sample contains paired video and action information. A virtual character moves through a 3D level while five cameras observe the character synchronously from different viewpoints. For every frame, we record the directional key state together with the relevant character and camera transforms. The action space contains nine discrete states:
\begin{equation}
\{\texttt{W},\texttt{A},\texttt{S},\texttt{D},\texttt{WA},\texttt{WD},\texttt{SA},\texttt{SD},\texttt{IDLE}\}.
\end{equation}

A standard trajectory contains 1,800 frames, corresponding to one minute of video, and produces five synchronized camera streams. The synthetic setting is valuable because every rendered visual transition has an explicitly known control signal rather than an inferred one.

Figure~\ref{fig:data-examples} shows representative rendered observations across different environments and camera viewpoints. Rather than showing only the five synchronized views from a single trajectory, we include examples across multiple scenes to illustrate both the visual diversity of the asset pool and the variation induced by the multi-view camera setup.

\subsection{Two-Stage Generation}
Due to the different execution requirements of real-time physics simulation and high-quality offline rendering, the pipeline performs trajectory generation and final rendering in two separate stages.

\paragraph{Stage I: physics-based trajectory generation.}
The engine enters \pie{} mode and runs an actual physics simulation. Synthetic directional inputs are sent to the character, while collision, gravity, slopes, and contact with geometry are resolved by Unreal Engine. No production-quality images are required at this stage. Instead, the pipeline records the character position, orientation, action state, and the information needed to construct camera trajectories.

\paragraph{Stage II: trajectory replay and offline rendering.}
A new Unreal Engine process reads the trajectory file. The character and cameras are placed at the recorded states frame by frame, and \mrq{} renders the sequence at high quality. Stage II does not resample actions or rely on a second physics simulation to determine motion.

\begin{figure}[t]
\centering
\includegraphics[width=\linewidth]{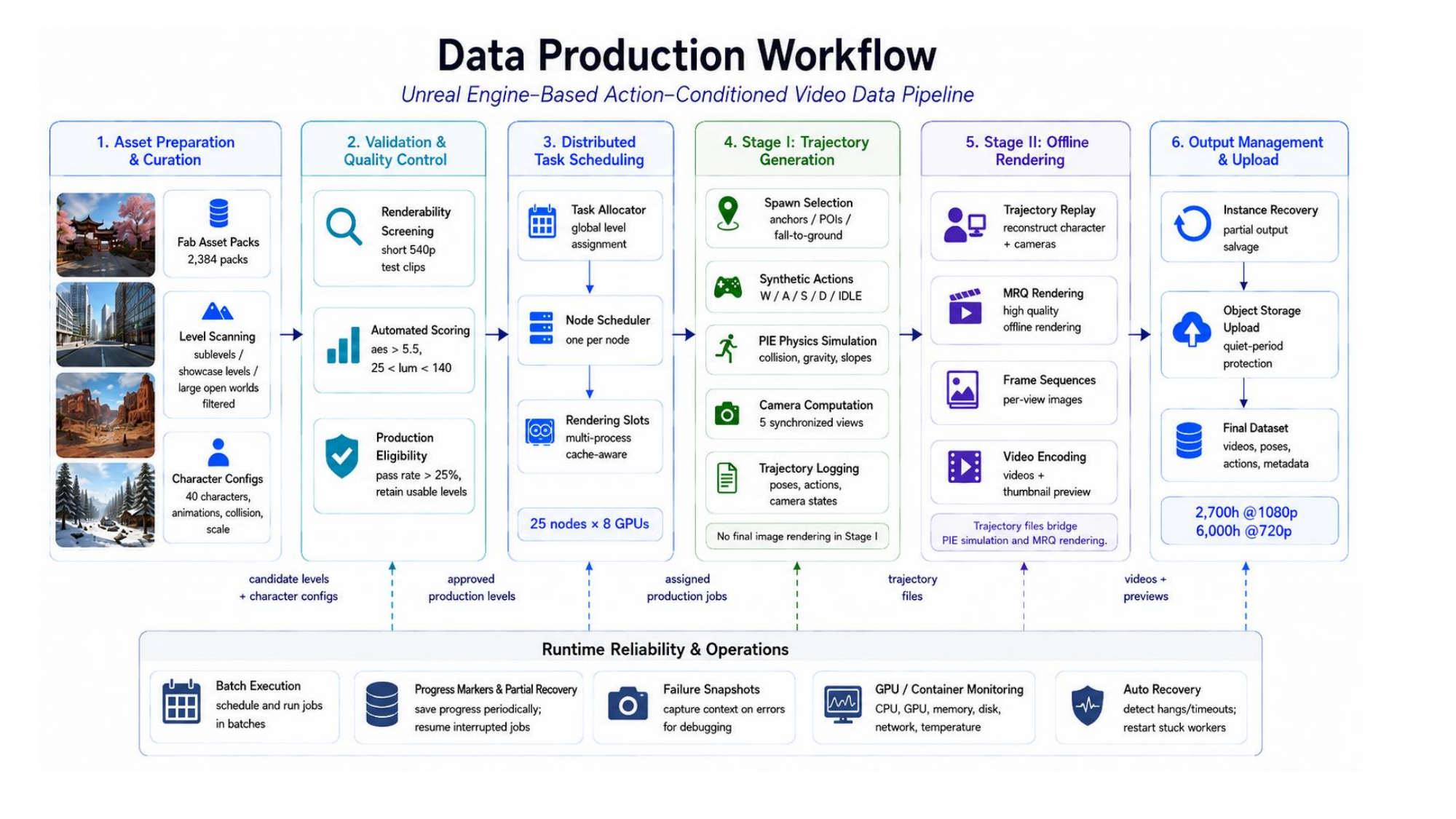}
\caption{End-to-end data production workflow. Stage I resolves motion under physics and writes an intermediate trajectory; Stage II replays the trajectory and performs high-quality offline rendering.}
\label{fig:pipeline-overview}
\end{figure}

This separation assigns a clear responsibility to each stage: {Stage I determines what happened; Stage II determines how well it is rendered.} The intermediate trajectory file acts as the contract between the two stages.
Figure~\ref{fig:pipeline-overview} summarizes how these components are integrated into the complete production workflow, from asset preparation and pre-production screening to distributed trajectory generation, offline rendering, and output management. The following sections describe the individual stages in detail.

\begin{figure}[t]
    \centering
    \begin{minipage}[c]{0.54\linewidth}
        \centering
        \vspace{2mm}

        \small
        \resizebox{\linewidth}{!}{
        \begin{tabular}{@{}p{0.68\linewidth}r@{}}
            \toprule
            Statistic & Value \\
            \midrule
            Purchased Fab asset packs & 2,384 \\
            Levels retained for production & 429 \\
            Humanoid characters & 40 \\
            Synchronized camera views & 5 \\
            Action states & 9 \\
            Frames per standard trajectory & 1,800 \\
            Standard trajectory duration & 60 s \\
            Compute nodes & 25 \\
            GPUs per node & 8 $\times$ RTX 5090 \\
            1080p output produced & 2,691 h \\
            720p output produced & 6,076 h \\
            Five-view production throughput & 33 min / node-hour \\
            Diagonal + backward + lateral action share & 46.6\% \\
            \bottomrule
            \end{tabular}
}
        \captionof{table}{Dataset and production statistics.}
        \label{tab:dataset-statistics}
    \end{minipage}
    \hfill
    \begin{minipage}[c]{0.42\linewidth}
        \vspace{0pt}
        \centering
        \includegraphics[width=\linewidth]{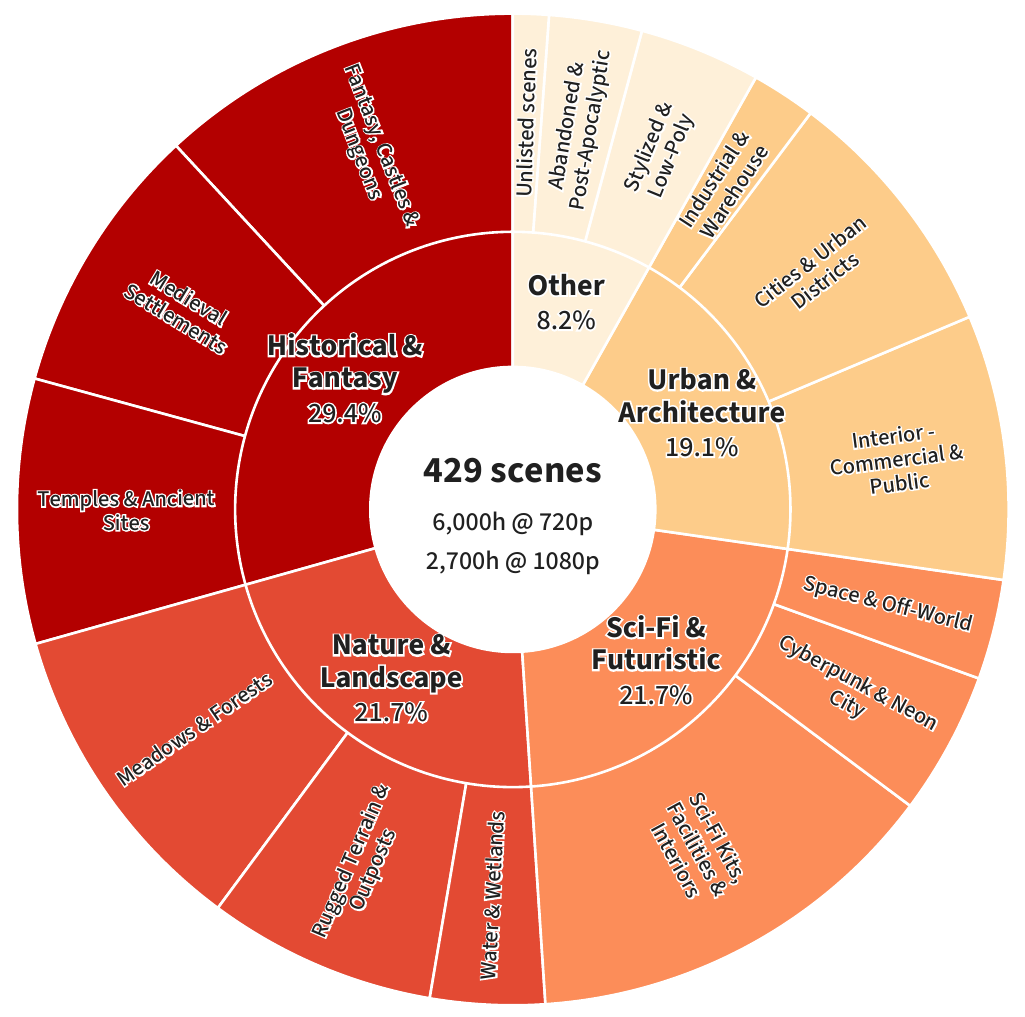}
        \captionof{figure}{
        Semantic diversity of the production scene pool. The inner and outer rings show major and fine-grained scene categories, respectively.
        }
        \label{fig:scene-diversity}
    \end{minipage}
\end{figure}

\section{Assets and Dataset Curation}

\subsection{Asset Preparation}

\paragraph{Scene assets and configuration.}
Scene assets are acquired in bulk from the Fab marketplace. Purchased packages are stored centrally in object storage and distributed to node-local disks on demand. We avoid loading scenes directly from object storage because Unreal Engine scene loading involves a large number of small random file accesses, for which network-storage latency becomes inefficient.
Asset packages do not follow a uniform directory structure. Some contain additional nesting layers, while others organize their levels and content differently. The pipeline therefore records the actual content directory of each package rather than assuming a fixed path convention. A global rendering configuration serves as the inventory of scene assets. For each asset pack, it records the asset identifier, content directory, candidate levels, and whether each level is enabled for production. The configuration also supports per-level parameter overrides, most commonly walking speed for spatially constrained scenes.
The configuration is generated automatically by scanning asset directories. During scanning, the pipeline excludes sub-levels that contain only partial scene content when opened independently, showcase levels that merely arrange models for asset preview, and very large open-world levels that rely on partitioned loading and are substantially more failure-prone in headless automation.

\paragraph{Character assets.}
Character assets require additional calibration because the raw mesh and skeleton do not provide all information required by the production pipeline. For each character, the configuration records the model path, animations for the nine action states, collision dimensions, model scale, orientation correction, walking speed, locomotion mode (walking, flying, or swimming), and an optional initial height offset.
Collision dimensions are estimated automatically from the model bounding box, scaled by a safety factor slightly larger than one, and clamped by a minimum size. Here we note that, during later extensions to non-humanoid assets, collision size and visual height cannot always be treated as the same quantity. For very flat characters, such as a bat with extended wings, the minimum collision size may still be physically reasonable but can severely overestimate the height used for camera framing. We therefore use the actual visual height, rather than the clamped collision dimension, when computing camera placement for such characters.

Character assets are stored once per node and shared across rendering slots through symbolic links. Scene assets are substantially larger and are therefore copied into a slot-specific Unreal project only while the corresponding task is active.

\subsection{Pre-Production Screening}

Before a level enters large-scale production, it passes through two stages of screening: a low-cost renderability check followed by automated visual-quality
evaluation.

\paragraph{Renderability screening.}
For each candidate level, we render five 10-second clips at 540p and manually inspect the results. Levels are rejected if they crash on load, contain no valid walkable region, cause persistent character hovering because of incorrect or invisible collision geometry, or produce substantial interpenetration between the character and scene geometry. Failing levels are disabled in the global configuration. Although this step requires manual inspection, its cost is small compared with allowing an invalid level to repeatedly occupy production slots while generating unusable data or no output at all.

\paragraph{Visual-quality filtering.}
Renderability alone does not guarantee useful visual data. We therefore apply an automated scoring procedure adapted from the SpatialVID scoring pipeline~\citep{wang2025spatialvid}. Each individual camera view is evaluated using two metrics: an aesthetic score (\texttt{aes}), produced by a pretrained image-quality model and intended to capture factors such as composition, color, and sharpness; and a luminance score (\texttt{lum}), based on average image brightness and used to reject excessively dark or overexposed samples.

Before full production, every renderable level is evaluated using twenty 10-second clips at 540p. A sampled clip passes the automatic filter if

\begin{equation}
    \texttt{aes} > 5.5,
    \qquad
    25 < \texttt{lum} < 140.
\end{equation}

We then compute a per-level pass rate and retain a level for production only if more than 25\% of its sampled outputs satisfy these criteria. The tiled five-view preview is excluded from scoring because its composite image statistics are not directly comparable with those of an individual camera view.

\subsection{Dataset Scale and Diversity}

Table~\ref{tab:dataset-statistics} summarizes the scale of the production effort, while Figure~\ref{fig:scene-diversity} characterizes the semantic diversity of the retained scene pool. The four largest semantic categories---Historical \& Fantasy, Sci-Fi \& Futuristic, Nature \& Landscape, and Urban \& Architecture---each account for approximately 19--29\% of the production scenes, indicating that the dataset is not dominated by a single scene type.

The action space covers all nine states. Diagonal, backward, and lateral movement together account for 46.6\% of the current action distribution, indicating that the generated trajectories are not dominated by simple forward motion.

\section{System Architecture}

\subsection{Distributed Production Architecture}
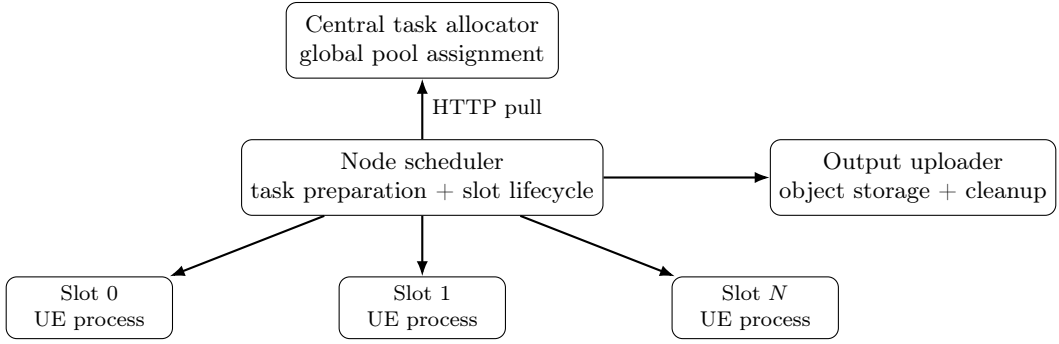
\begin{figure}[t]
\centering
\begin{tikzpicture}[
    node distance=8mm and 9mm,
    box/.style={draw, rounded corners, align=center, minimum width=36mm, minimum height=10mm, font=\small},
    smallbox/.style={draw, rounded corners, align=center, minimum width=22mm, minimum height=8mm, font=\footnotesize},
    arrow/.style={-{Latex[length=2mm]}, thick}
]
\node[box] (allocator) {Central task allocator\\global pool assignment};
\node[box, below=of allocator] (scheduler) {Node scheduler\\task preparation + slot lifecycle};
\node[smallbox, below left=of scheduler] (s0) {Slot 0\\UE process};
\node[smallbox, below=of scheduler] (s1) {Slot 1\\UE process};
\node[smallbox, below right=of scheduler] (sn) {Slot $N$\\UE process};
\node[box, right=22mm of scheduler] (uploader) {Output uploader\\object storage + cleanup};

\draw[arrow] (scheduler) -- node[right,font=\footnotesize]{HTTP pull} (allocator);
\draw[arrow] (scheduler) -- (s0);
\draw[arrow] (scheduler) -- (s1);
\draw[arrow] (scheduler) -- (sn);
\draw[arrow] (scheduler) -- (uploader);
\end{tikzpicture}
\caption{Distributed production architecture. The central service performs coarse-grained assignment, while each node independently manages high-frequency slot lifecycle operations and output transfer.}
\label{fig:distributed-architecture}
\end{figure}

Large-scale operation involves three types of long-running processes: a central {task allocator}, one {node scheduler} per compute node, and one independent {output uploader} per node. Figure~\ref{fig:distributed-architecture} summarizes their responsibilities.

The {task allocator} maintains global level-pool assignments and serves tasks to nodes over HTTP, but does not launch or monitor individual Unreal Engine processes. The {node scheduler} manages execution locally, including asset preparation, Stage I and Stage II process launches, stall detection, retries, and cleanup. The {output uploader} independently transfers completed outputs to object storage and removes local copies only after successful upload, preventing network I/O from blocking rendering.

Keeping high-frequency execution management local avoids routing process-level status from hundreds of rendering slots through a central service. The allocator therefore handles only coarse-grained global assignment, while each node maintains its own execution state.

\subsection{Cache-Aware Task Scheduling}
A naive scheduler could let every node sample uniformly from the global level list. In practice, this performs poorly because Unreal Engine builds substantial scene-specific local state: shaders are compiled, acceleration structures are constructed, and textures are cached. The first render of a scene can therefore incur a cold-start delay of more than ten minutes, whereas subsequent renders can reuse local caches.

To preserve locality, we partition the level list into a number of pools equal to the number of nodes. Each node is persistently bound to one pool and repeatedly renders only scenes from that subset. After the pool is exhausted, its order is reshuffled and another pass begins. Since trajectory initialization is randomized, repeated rendering of the same level still produces distinct data.

Pool assignments are persisted to disk. If assignments were regenerated after allocator restart, nodes would switch to new scene subsets and discard the benefit of their accumulated caches. In production, pool-based scheduling reduces per-scene rendering time to less than half of cold-start rendering time for repeatedly visited scenes.

\subsection{Slot-Level Execution}
Each node contains eight GPUs and runs multiple rendering slots. A slot corresponds to an independent Unreal Engine process lifecycle. More than one process may share a GPU because a single process does not continuously saturate the device; significant time is spent loading assets, compiling shaders, or waiting on disk I/O. The optimal concurrency factor is tuned empirically because excessive concurrency causes memory pressure and process interference.

A slot follows a fixed lifecycle:

\begin{center}
\small
\textbf{task assignment}
$\rightarrow$
\textbf{asset preparation}
$\rightarrow$
\textbf{trajectory generation} (Stage I)
$\rightarrow$
\textbf{rendering} (Stage II)
$\rightarrow$
\textbf{cleanup}.
\end{center}

After receiving a level from the allocator, the scheduler copies the required scene assets into the slot-specific Unreal project and launches Stage I to generate trajectory files. Once Stage I terminates, the scheduler determines which trajectories were successfully produced and launches Stage II only for those trajectories. After rendering completes, the copied scene assets are removed and the slot returns to the idle pool.

Scene assets are copied on demand because Unreal Engine expects content to be available within the project directory. Keeping the full scene collection resident in every slot would require excessive disk capacity and would substantially increase project-scanning time at engine startup. Furthermore, we note that headless execution requires a virtual display environment; each slot is assigned an independent virtual display identifier so that concurrent Unreal Engine processes do not interfere with one another.

\section{Trajectory Generation}

\subsection{Stage-I Execution and Initialization}

Stage I executes control logic over thousands of Unreal Engine frames without manual interaction. Although Unreal Engine can launch a Python script at startup, trajectory generation cannot be implemented as a conventional blocking script because the engine must continue advancing between control steps.

\paragraph{Frame-driven control.} The pipeline therefore registers a \textbf{per-frame callback}. The startup script performs initialization and installs a function that is invoked on every engine frame. At each invocation, the callback performs only the operation required by the current state---for example, checking whether the level has loaded, issuing the current action, reading the character transform, or transitioning to the next state---and immediately returns control to the engine.

The resulting automation is implemented as a frame-driven state machine. Conditions such as waiting for scene initialization are evaluated incrementally across frames rather than through blocking loops, which would prevent Unreal Engine from advancing. The Stage I state machine can be summarized as:

\begin{center}
\small
initialize $\rightarrow$ enter runtime $\rightarrow$ prepare scene/character
$\rightarrow$ warm up $\rightarrow$ trajectory loop $\leftrightarrow$ respawn
$\rightarrow$ exit runtime $\rightarrow$ compute cameras
$\rightarrow$ next trajectory.
\end{center}

During trajectory execution, actions are applied through Unreal Engine's runtime simulation, and the resulting character transform is read back at every frame. The control code does not analytically integrate character motion; collision, gravity, landing, and slope traversal are therefore resolved by the engine itself.

\paragraph{Character preparation and spawning.} At the beginning of each trajectory, a character is sampled and instantiated using its configured model, animation mappings, scale, orientation correction, and collision dimensions. Character models and animations are cached within a task to avoid redundant loading.

Trajectory quality is sensitive to the initial position. The pipeline first uses level-provided player starts or camera anchors when suitable locations are available. Otherwise, it samples positions near large scene objects, which serve as approximate points of interest. After selecting a horizontal position, the character is placed above the candidate point and allowed to fall under engine physics until stable ground contact is reached. Candidates that fail to settle within a time limit are rejected and resampled. Flying and swimming characters use configured height offsets and gravity settings appropriate to their locomotion mode.

\subsection{Action Sampling and Recovery}

At each action decision point, one of the nine discrete action states is sampled from a predefined probability distribution and held for a randomly sampled number of frames before the next action is selected. Temporal holding prevents the high-frequency reversals and visible jitter that arise when actions are resampled independently at every frame. The duration distribution therefore controls a trade-off between locally coherent motion and the frequency of action transitions.

\paragraph{Point-of-interest guidance.}
Purely random motion can leave a character exploring only a limited region near its initial position. To encourage broader spatial coverage without changing the action distribution, the pipeline gradually adjusts the character's facing direction toward a sampled point of interest while leaving action-sampling probabilities unchanged.

\paragraph{Dead-end recovery.}
The pipeline detects cases in which the character receives nonzero movement input but exhibits negligible displacement over multiple frames. Once the condition persists beyond a threshold, an escape maneuver rotates the character by a substantial angle before motion continues. If repeated escape attempts fail, the local position is abandoned and the character is respawned elsewhere.

\subsection{Multi-View Camera Trajectories}

After the character trajectory has been fixed, Stage I computes five synchronized camera trajectories: a back follow view, left-side and right-side views, an elevated top-down view, and a first-person view near the character's head, as shown in Figure~\ref{fig:data-examples}. The four external cameras are defined through character-relative offsets, so their desired world-space positions evolve with the character orientation.

\paragraph{Occlusion handling.} A fixed relative offset can place an external camera behind or inside scene geometry. For each frame, the pipeline casts a ray from the character toward the desired camera position. When an obstacle is detected, the camera is moved inward to remain in front of the obstructing geometry. Directly applying this correction can produce visible snapping when narrow objects temporarily occlude the desired camera position. Camera distance is therefore smoothed asymmetrically: it contracts quickly when an obstacle appears to prevent geometry penetration, but expands more slowly after the obstacle disappears to reduce visual popping. Because camera trajectories are computed after the physical character motion is complete, occlusion queries and camera corrections operate on a fixed motion sequence without affecting the underlying trajectory.

\section{Offline Rendering}

\subsection{Why MRQ?}

Our initial implementation captured frames directly during Stage I using
Unreal Engine \texttt{SceneCapture} components. This avoided trajectory replay and allowed simulation and image capture to occur in the same process. In our scripted production setup, however, we were unable to obtain sufficiently stable high-quality anti-aliasing despite testing multiple control paths and console-variable configurations, and visible aliasing remained in the output.

We therefore moved final image generation to Unreal Engine's Movie Render Queue(MRQ), the engine's offline rendering pipeline for high-quality linear content. Unlike real-time capture, MRQ is not constrained by an interactive frame-time budget and can spend substantially more computation on each output frame. In particular, it supports spatial and temporal supersampling, in which multiple render evaluations are accumulated into a single output frame. Spatial samples evaluate the same moment with sub-pixel sampling offsets, while temporal samples evaluate multiple sub-frame times within the output-frame interval. These mechanisms provide substantially better anti-aliasing and temporal rendering quality than we obtained from our Stage-I capture setup.

MRQ also exposes rendering-specific controls that are useful for unattended high-quality production, including configurable warm-up, reusable render settings, console-variable overrides, and image-sequence output. In our pipeline, the primary benefit is the ability to prioritize image quality over real-time execution and to produce stable frame sequences for subsequent video encoding.

This rendering mode is not naturally compatible with the real-time physics execution used during Stage I. Stage I must continuously advance the world to resolve character motion, collision, and gravity, whereas MRQ controls timeline evaluation according to its offline rendering process and may evaluate an output frame multiple times. Under these constraints, we first record the realized physical trajectory and then reconstruct it as a deterministic timeline for MRQ. Stage II therefore renders previously determined motion rather than running the character controller and physics simulation again.

\subsection{MRQ-Based Rendering Workflow}

Once a Stage-I trajectory has been completed, Stage II reconstructs the recorded character and camera states as an Unreal Engine timeline that can be evaluated by MRQ. For each trajectory, the pipeline creates one character object and five camera objects and uses the transforms recorded in Stage I as the source of truth. Stage II does not resample actions or rerun the character controller and physics simulation; it replays the realized motion directly. After the timeline and level are prepared, MRQ renders the five camera streams as image sequences, which are subsequently encoded into video.

\paragraph{Transform reconstruction.} Each transform track contains translation, rotation, and scale channels. Although scale is constant for most trajectories, it must still be written explicitly. In an early implementation, only translation and rotation were populated under the assumption that the character would retain its configured scale. Timeline evaluation instead reset the scale, producing incorrectly sized characters. The replay pipeline therefore writes all transform components explicitly. Additionally, vertical placement requires additional care. During Stage I, the recorded character position corresponds to the center of the collision capsule rather than directly to the visible mesh origin. Moreover, Unreal Engine may internally constrain the effective capsule dimensions relative to the values requested by the configuration. Stage I therefore records the actual capsule height used by the engine, and Stage II uses this observed value when reconstructing the vertical mesh offset. This avoids systematic floating or sinking of characters during replay.

\paragraph{Level sanitization.} Third-party asset packs may contain interactive or cinematic behavior that is appropriate for demonstration scenes but conflicts with automated rendering. In particular, some levels automatically play built-in cinematic sequences after loading. These sequences can take control of the active camera and cause the rendered output to switch unexpectedly from the pipeline-controlled view to a showcase camera. Before trajectory replay begins, the pipeline scans the level and removes such sequence actors so that camera control remains with the generated trajectory.

\paragraph{Texture residency.} Texture streaming introduces a second rendering-specific issue. Unreal Engine normally loads texture detail progressively in order to preserve interactive performance. During offline rendering, however, this behavior can cause early frames to contain visibly lower-resolution textures before higher-resolution mip levels become resident. Before MRQ rendering starts, the pipeline therefore adjusts texture-streaming settings to favor visual consistency over interactive memory efficiency. It increases the available texture pool, encourages relevant textures to remain fully resident, relaxes normal streaming constraints, and biases texture detail toward higher-resolution levels. These settings reduce the risk of blurred textures during the initial portion of a rendered trajectory.

\paragraph{Frame output and video encoding.} MRQ produces image sequences for the five camera views. After rendering completes, the pipeline encodes each view into a separate video and generates an additional tiled five-view preview for efficient inspection. The raw image sequences are deleted after successful encoding because they require roughly an order of magnitude more storage than the encoded videos and can rapidly exhaust node-local disk capacity.

\section{Reliability and Operations}

\begin{figure*}[t]
\centering
\includegraphics[width=0.8\textwidth]{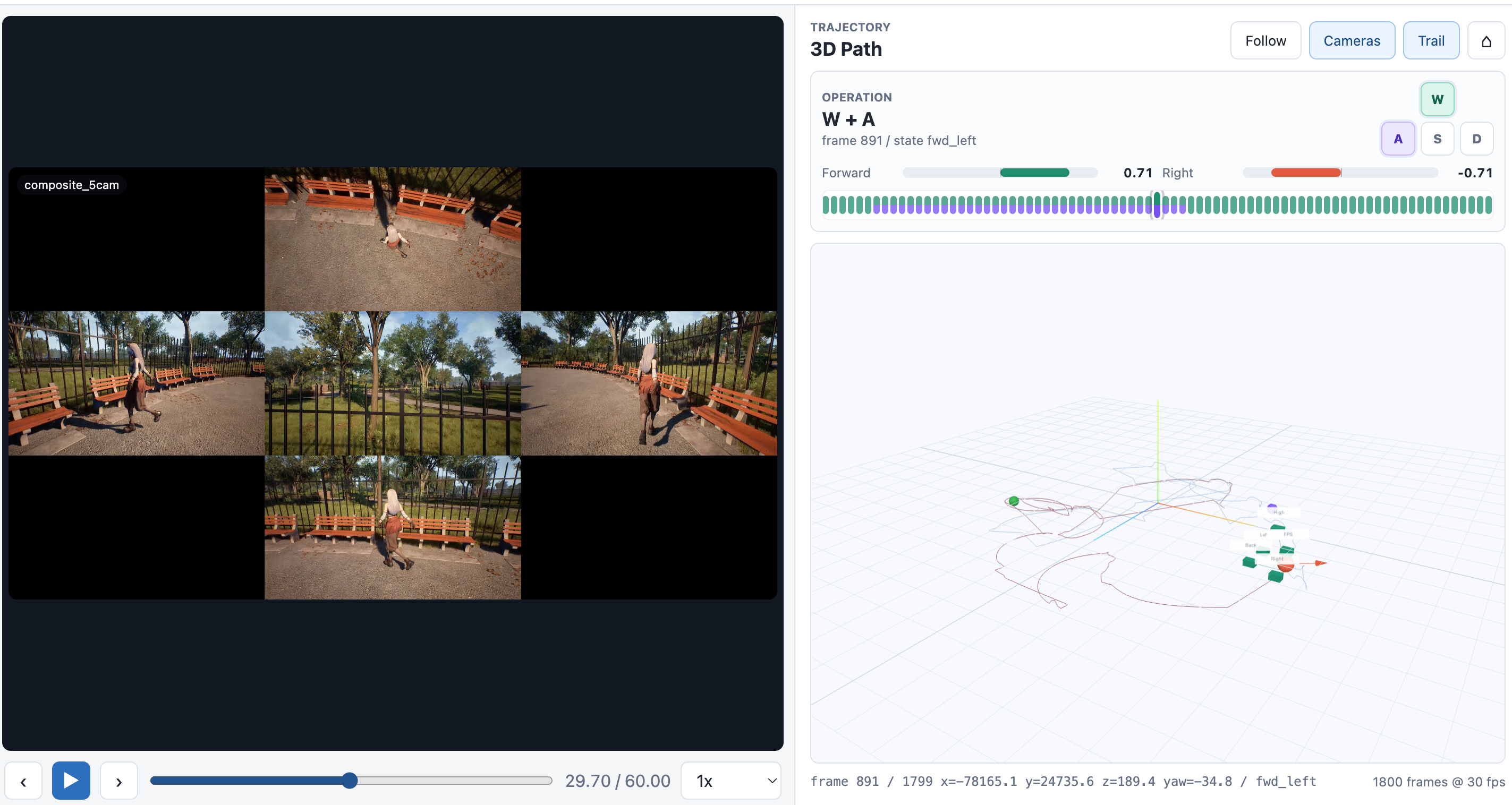}
\caption{
Trajectory inspection interface used for qualitative verification and production debugging. The left panel shows the synchronized multi-view video, while the right panel displays the frame-level action state together with the 3D character and camera trajectories.
}
\label{fig:trajectory-inspector}
\end{figure*}

\subsection{Output Organization}
Outputs are organized hierarchically as
\begin{center}
\texttt{scene / level / instance / trajectory}.
\end{center}
Each instance corresponds to one execution of a production task and contains the generated trajectories together with their videos, action records, camera states, and execution metadata. The instance identifier combines node identity, timestamp, and a local sequence number, preventing concurrent or repeated processing of the same level from overwriting existing outputs.

For qualitative inspection and debugging, we additionally design a lightweight visualization interface for completed trajectories. Figure~\ref{fig:trajectory-inspector} shows an example. The interface synchronizes the rendered multi-view video with the corresponding frame-level action state and visualizes the recorded character and camera trajectories in 3D space. Users can scrub through the trajectory frame by frame while observing the active action, character motion, and relative camera positions. The viewer is not part of the automatic acceptance or recovery logic. Instead, it is used for spot-checking generated data and diagnosing issues that are difficult to identify from logs alone, such as unexpected camera motion, incorrect action alignment, abnormal character movement, or trajectories that remain confined to a small region.

\subsection{Completion Detection and Partial Recovery}
\label{sec:completion-detection}
The scheduler and Unreal Engine rendering process are independent. Unreal Engine exposes no convenient external query interface for the scheduler, and the engine may crash at any time. We therefore use an append-only progress log as a one-way communication channel. After each trajectory completes, the engine writes a marker in a fixed format. The scheduler counts these markers to determine progress. No bidirectional handshake is required: the renderer writes and the scheduler reads. This loose coupling is robust to transient failures on either side. A separate instance-level completion marker is written only after all trajectories finish and the engine completes normal shutdown. Its absence indicates that the instance was interrupted and must be examined for partial recovery. If an instance crashes after completing some trajectories, we preserve any trajectory for which both a completion marker and the expected video files exist. Incomplete trajectories are deleted. If no valid trajectory remains, the instance is discarded. This design assumes failure can occur at arbitrary points rather than treating partial execution as exceptional.

\subsection{Asynchronous Upload and Failure Snapshots}
The uploader runs independently from rendering. It scans completed local outputs, uploads them to object storage, and deletes local copies only after successful transfer. A {quiet-period rule} prevents the uploader from touching directories that were modified recently. This avoids racing with a video encoder that is still writing files. When a rendering task fails, the scheduler packages the failure context into a dedicated snapshot containing engine logs, exit code, failure reason, slot and GPU identifiers, and progress information. Failure snapshots are uploaded separately because local logs may be removed during cleanup and would otherwise be unavailable for postmortem analysis.

\subsection{Cluster Startup and Monitoring}
A cluster management script supports start, stop, status inspection, and health checks. On startup, it verifies node reachability, container availability, and whether a scheduler is already running before launching a new one. The scheduler must run under a designated non-privileged user. Starting it with administrative privileges causes Unreal Engine processes on the node to terminate within seconds and leaves difficult-to-clean temporary files. 

Production progress is tracked through three complementary signals. Scheduler logs provide the freshest indication of recently completed trajectories; object-storage listings provide the authoritative count of delivered outputs but lag behind production because of upload queueing and quiet-period protection; and allocator state is primarily used to inspect task-pool balance and assignment behavior rather than throughput.

Two central monitoring loops handle GPU-driver access failures and disk pressure. Monitoring runs outside the production nodes because a node suffering extreme load, a full disk, or a graphics failure may not be able to monitor itself reliably. One recurring GPU failure mode is characterized by the container losing functional GPU access while the host still reports the devices normally. Rebuilding the container restores device mappings, after which the scheduler is relaunched. Recovery scripts use file locks to prevent overlapping runs and cooldown periods to avoid repeated rebuilds of the same node. 

Representative diagnosed failure cases across the engine, container, hardware,
and storage layers are summarized in Appendix~\ref{app:failure-modes}.

\section{Limitations and Discussion}

\paragraph{Data-Production Trade-offs}
The system is not optimized for cinematic visual quality alone. The target data must balance several competing goals: \textbf{controllability}, \textbf{action diversity}, \textbf{physical plausibility}, \textbf{visual quality}, and \textbf{production scalability}. These objectives can conflict. Stronger point-of-interest guidance may produce more purposeful-looking motion but reduce action diversity. Stricter aesthetic filtering may increase average visual quality while discarding structurally useful environments. More camera views increase observation diversity but raise rendering and storage cost. The design choices in this report should therefore be understood as compromises for large-scale action-conditioned data production rather than globally optimal choices for every downstream task.

\paragraph{Perceptual Quality and Downstream Utility}
The current curation pipeline relies on perceptual proxies, primarily aesthetic and luminance scores. These metrics are useful for removing obvious failures, including severely dark, overexposed, visually empty, or poorly rendered samples. However, perceptual quality is not necessarily equivalent to utility for learning environment dynamics. A visually unattractive scene may still contain useful geometry, motion, or interaction structure. The current thresholds were selected empirically and have not yet been calibrated against downstream world-model performance. An open question is therefore how perceptual filtering affects learned dynamics: overly aggressive filtering may discard useful structural diversity, whereas permissive filtering may retain low-quality samples that interfere with training. Establishing this relationship requires controlled downstream experiments and remains outside the scope of the present production-focused report.

\paragraph{Action Semantics and Future Extensions}
The current report primarily describes a character-centric control scheme in which directional inputs are defined relative to the character. A newer implementation moves toward a {single fixed camera with actions defined in camera coordinates}. Under this representation, pressing the forward key means moving deeper into the image rather than moving in the direction the character currently faces. The character turns toward its movement direction, and obstacle recovery can rotate the camera around the character rather than rotating the character in place.

Character-centric and camera-centric controls represent two different choices of action coordinates rather than a strict quality trade-off. The former defines movement relative to the character orientation, while the latter defines movement relative to the camera view. Both conventions are widely used in interactive systems. We are exploring the camera-centric variant to study how the choice of action coordinates affects action-conditioned prediction, rather than assuming that either representation is intrinsically superior.

\paragraph{Limited action repertoire.}
The current action space is centered on locomotion: forward, backward, lateral and diagonal movement, turning behavior, and idle periods. It does not yet include richer actions such as jumping, running, crawling, attacking, or complex environmental interactions such as climbing over obstacles.

\section{Conclusion}
We presented a large-scale Unreal Engine pipeline for producing multi-view, action-conditioned video for world-model pretraining. The pipeline converts heterogeneous third-party 3D assets into synchronized video, action, character, and camera trajectories while preserving physical motion generated by the game engine. To accommodate the different execution requirements of real-time physics simulation and high-quality MRQ rendering, trajectory generation and final rendering are performed in separate stages connected through recorded per-frame states.

Operating this pipeline at cluster scale requires substantially more than rendering alone. The production system integrates asset discovery and screening, cache-aware distributed scheduling, node-local rendering slots, trajectory and camera generation, MRQ-based offline rendering, asynchronous output transfer, partial recovery, and cluster-level monitoring. Across 25 nodes with eight RTX 5090 GPUs each, the system retains 429 production levels from 2,384 acquired asset packs and supports 40 characters and five synchronized camera views. It has produced approximately 2,700 hours of 1080p video and 6,000 hours of 720p video.

The report focuses on the infrastructure required to make such data production scalable and operationally robust rather than on demonstrating improvements to a particular world-model architecture. An important next step is to connect production decisions more directly to downstream learning outcomes, including how scene selection, perceptual filtering, action semantics, and trajectory distributions affect action-conditioned prediction. We hope that documenting the production pipeline and its operational constraints provides a useful reference for future large-scale synthetic-data systems for interactive world models.

\bibliographystyle{plainnat}
\bibliography{world_model_data_pipeline_refs}

\beginappendix

\section{Key Configuration Parameters}
The configurable parameters fall into three groups.

\paragraph{Rendering configuration.}
Asset storage location and content directory, level path inside Unreal Engine, production enable/disable flag, available trajectory count, and optional per-level overrides such as walking speed and exposure compensation.

\paragraph{Character configuration.}
Model path, animation paths for the nine action states, collision volume, model scale, orientation correction, walking speed, locomotion mode, and initial height offset.

\paragraph{Scheduler parameters.}
GPU list, number of processes per GPU, output directory, rendering resolution and frame count, number of trajectories, upload destination, disk-space threshold, retry count, allocator address, and node identifier.

\section{Directory Conventions}
\begin{itemize}[leftmargin=1.6em]
    \item \textbf{Scene assets:} stored in a node-local shared directory and copied into slot-specific project directories on demand.
    \item \textbf{Character assets:} stored once per node and shared across slots by symbolic links.
    \item \textbf{Trajectory intermediates:} produced by Stage I and removable after Stage II consumes them.
    \item \textbf{Rendered outputs:} organized as \texttt{scene / level / instance / trajectory}.
    \item \textbf{Failure snapshots:} stored separately and used for postmortem analysis.
\end{itemize}

\section{Common Troubleshooting Procedure}
When diagnosing production failures, we inspect the system in the following order:
\begin{enumerate}[leftmargin=1.6em]
    \item \textbf{Scheduler log}: active slots, success/failure counters, and disk capacity.
    \item \textbf{Unreal Engine log}: crash location or latest progress point.
    \item \textbf{Progress log}: number of completed trajectories in a specific instance.
    \item \textbf{GPU state inside the container and on the host}: both views are required to diagnose stale container device mappings.
    \item \textbf{Failure snapshot}: postmortem analysis and aggregation of failure categories.
\end{enumerate}

\section{Representative Production Failure Cases}
\label{app:failure-modes}

Large-scale production exposed failures across multiple layers of the system, including Unreal Engine execution, container state, GPU drivers, and local storage. Table~\ref{tab:failure-modes} summarizes representative cases that were diagnosed during production.

The mapping between an observed symptom and its cause is case-specific; similar high-level symptoms may arise from different underlying failures. We therefore report the diagnostic evidence used in each case together with the mitigation that proved effective in our production environment.

\begin{table*}[t]
\centering
\small
\setlength{\tabcolsep}{5pt}
\renewcommand{\arraystretch}{1.15}

\caption{Representative diagnosed production failures and their case-specific causes.}
\label{tab:failure-modes}

\begin{tabularx}{\textwidth}{
    >{\raggedright\arraybackslash}p{0.20\textwidth}
    >{\raggedright\arraybackslash}X
    >{\raggedright\arraybackslash}X
    >{\raggedright\arraybackslash}X
}
\toprule
Observed failure & Diagnostic signal & Identified cause & Mitigation \\
\midrule

\multicolumn{4}{l}{\textbf{Engine}} \\
\addlinespace[2pt]

Crash after switching character models
& Engine log reports a skeletal assertion
& Residual skeletal state from the previously loaded character is incompatible with the new model
& Explicitly clear the previous model state before loading the next character \\

Crash while loading a large partitioned level
& Stack trace points to the world-partition loading subsystem
& Partitioned-world state is accessed before initialization has completed
& Exclude the level from automated production \\

Rendered viewpoint unexpectedly switches to a showcase camera
& Video changes from the pipeline-controlled camera to a fixed cinematic view
& An auto-playing sequence actor takes ownership of the active camera
& Remove or disable built-in sequence actors before rendering \\

Character moves through the scene while the walking animation appears stationary or inconsistent
& Actor transform changes normally, but expected leg motion is absent
& Animation root motion conflicts with externally driven character transforms
& Disable or lock root-bone translation for the affected animation setup \\

\midrule
\multicolumn{4}{l}{\textbf{Container}} \\
\addlinespace[2pt]

Unreal Engine fails to start because the GPU is unavailable inside the container
& GPU queries fail inside the container while the host still reports the device normally
& Container GPU device mapping has become stale or invalid
& Rebuild the container and restart the scheduler \\

Containerized rendering process exits under high memory usage
& Exit status and runtime logs indicate an out-of-memory event
& The configured container memory limit is exceeded
& Inspect slot concurrency and memory usage, then restart or rebuild the workload \\

\midrule
\multicolumn{4}{l}{\textbf{Hardware}} \\
\addlinespace[2pt]

Multiple GPU processes become unresponsive at the node level
& Kernel or driver error signals appear and affected processes enter uninterruptible wait
& GPU or driver-level failure prevents normal process recovery
& Hard reboot the affected node \\

\midrule
\multicolumn{4}{l}{\textbf{Storage}} \\
\addlinespace[2pt]

Scene assets remain in rendering slots after an interrupted scheduler run
& Slot directories still contain copied scene packages after production has stopped
& Forced termination bypassed the normal cleanup path
& After confirming that no Unreal Engine process is using the files, clean the slot before restart \\

System disk usage grows rapidly during repeated task failures
& Log directories expand continuously while the same failure recurs
& Persistent engine or container logging accumulates faster than cleanup can reclaim space
& Reclaim accumulated logs and eliminate the recurring failure source \\

\bottomrule
\end{tabularx}
\end{table*}

\end{document}